# DECO-MWE: Building a Linguistic Resource of Korean Multiword Expressions for Feature-Based Sentiment Analysis

**Jaeho Han, Changhoe Hwang, Seongyong Choi, Gwanghoon Yoo, Eric Laporte*& Jeesun Nam**
DICORA, Department of Linguistics and Cognitive Science,
Hankuk University of Foreign Studies, Korea
* Université Paris-Est, LIGM, CNRS, UPEM, ESIEE, ENPC, France
hanjaeho0308@gmail.com, hch8357@naver.com, csy@hufs.ac.kr, rhkdgns2008@naver.com, namjs@hufs.ac.kr
*eric.laporte@univ-paris-est.fr

**Abstract**

This paper aims to construct a linguistic resource of Korean Multiword Expressions for Feature-Based Sentiment Analysis (FBSA): DECO-MWE. Dealing with multiword expressions (MWEs) has been a critical issue in FBSA since many constructs reveal lexical idiosyncrasy. To construct linguistic resources of sentiment MWEs efficiently, we utilize the Local Grammar Graph (LGG) methodology: DECO-MWE is formalized as a Finite-State Transducer that represents lexical-syntactic restrictions on MWEs. In this study, we built a corpus of cosmetics review texts, which show particularly frequent occurrences of MWEs. Based on an empirical examination of the corpus, four types of MWEs have been discerned. The DECO-MWE thus covers the following four categories: Standard Polarity MWEs (SMWEs), Domain-Dependent Polarity MWEs (DMWEs), Compound Named Entity MWEs (EMWEs) and Compound Feature MWEs (FMWEs). The retrieval performance of the DECO-MWE shows 0.806 f-measure in the test corpus. This study brings a two-fold outcome: first, a sizeable general-purpose polarity MWE lexicon, which may be broadly used in FBSA; second, a finite-state methodology adopted in this study to treat domain-dependent MWEs such as idiosyncratic polarity expressions, named entity expressions or feature expressions, and which may be reused in describing linguistic properties of other corpus domains.

## 1. Introduction

This study presents a linguistic resource of Korean Multiword Expressions for Feature-Based Sentiment Analysis (FBSA): DECO-MWE. A Recursive Transition Network methodology called Local-Grammar Graphs (Gross 1997, 1999) is adapted to construct the resources: they are compiled into Finite State Automata and Finite State Transducers and coupled with the DECO Korean Electronic dictionary that provides a diversity of linguistic information such as morphological, syntactic, semantic and sentiment-polarity information (Nam 2012, 2015). DECO-MWE covers 4 types of MWEs: Standard Polarity MWEs (SMWEs), Domain-dependent Polarity MWEs (DMWEs), Compound Named Entity MWEs (EMWEs) and Compound Feature MWEs (FMWEs). Due to the difficulties in processing the idiosyncrasy of MWEs, MWEs need to be empirically described in resources effectively structured for automatic processing. Moreover, since the Korean language shows extremely complex morphological characteristics, the resources should reliably recognize all inflectional variations of MWEs. This paper will discuss an effective way to build a linguistic resource of Korean MWEs for FBSA.

Sentiment Analysis (SA) mainly focuses on the classification of Semantic Orientation (SO), commonly referred to as polarity. Two general approaches to SA are known: the lexicon-based approach that computes the polarity value of texts through sentiment lexicons containing polarity values, and the machine-learning approach that classifies polarity by mathematical algorithms trained on a sentiment dataset (Liu 2015). With lexicon-based methods, the performance of SA fundamentally depends on the quality and size of sentiment lexicons because this approach is deeply grounded in the keyword-based vectorization of sentiment-related vocabulary (Kim et al., 2009). As compared to document- or sentence-level classification, it has become apparent that Feature-based Sentiment Analysis performs finer-grained analysis (Liu 2012).

Most of current studies on SA are grounded in machine-learning approaches such as maximum entropy, SVM or Naïve Bayes classification (Pang et al., 2002). However, conventional methods show severe limitations for FBSA, which requires processing lexical and syntactic properties. In case of inflectional linguistic phenomena in the data, frequencies of MWEs carrying SO should be considered as well. Lexicon-based approaches are more suitable to deal with a variety of MWEs for FBSA since they make it possible to analyze and calculate lexical information on sentiment words, named entities and feature nouns.

For FBSA, Liu (2012) introduces the Sentiment Quintuple model, consisting of an Entity (e), Aspect (a), Sentiment Value (s), Opinion Holder (h), and Time (t). The FBSA approach to sentiment computation requires a sentiment-annotated corpus, annotated at the level of tokens. (Hu and Liu 2006). This corpus provides a diversity of sentiment-related information and determines the reliability of the analysis. In the case of single-word expressions, explicit sentiment values can be assigned with a standard dictionary, but MWEs pose complex problems in the implementation of FBSA, since the process cannot rely on the compositionality of single words' senses. For example, *buy something for a song* means ‘buy something for a low price’ and is unrelated to singing, which manifests the idiosyncrasy of MWEs. Another type of MWEs includes compound nouns such as *anti-aging cream* that should be properly recognized for the correct analysis of the target of evaluation. Such MWEs should not be processed as several words but tokenized as one unit, and this one should be placed in the same category as *cream* in FBSA. Therefore, MWEs should be properly chunked, tagged and lemmatized in the tokenizing phase for a reliable FBSA.

The following examples in Korean show MWEs expressing polarity (i.e. (a) (b)) and analyzable as named entities (i.e. (c) and (d)):

(1a) 눈길을 끌다/ *nwunkil(eyes)-eul kkeulta(attract)*
"to attract one's attention"

(1b) 마음이 가다/ *maeum(mind)-i kata (go)*
"to catch one's fancy"

(1c) 모이스쳐라이징 크림/*moiseuchyeolaicing kheulim*
"moisturizing cream"

(1d) 컬픽스 마스카라/ *kheolpikseu maseukhala*
"curl fix mascara"

'눈길/*nwunkil'* in (a) means 'eyes', and '마음/*maeum'* in (b) 'mind'. In (a) and (b), the expressions are about attracting attention and interest respectively, not moving eyes or mind. Furthermore, (c) indicates a 'cream' the function of which is 'moisturizing' and (d) a product 'mascara' having a curl-fixing function.

As mentioned above, they need to be chunked and tagged as a single unit each and associated with all inflectional variations in order to be properly analyzed. Given that Korean is an agglutinative language in which a word is generally composed of several morphemes, this aspect should be carefully considered in case of MWEs, which makes the processing of Korean MWEs more difficult than that of English ones.

Nonetheless, as is the case with English, the majority of Korean compound words are right-headed, exhibiting 'Modifier-Head' structure (Bejček and Pavel 2010). Therefore, nominal MWEs should be grouped into semantic categories according to their heads. For example, '래스트 파운데이션/ *layseutheu phawunteisyeon'* (long-lasting foundation) refers to some kind of '파운데이션/ *phawunteisyeon*' (foundation). Besides, among Named entity MWEs, English loanwords frequently occur, which disturbs an effective recognition of MWEs. Let us consider:

(2a) 래스트 파운데이션/ *layseutheu phawunteisyeon*

(2b) 라스트 파운데이션/ *laseutheu phawunteisyeon*

(2c) 라스트 파데/ *laeseutheu phatei*

Since the transliteration of English loanwords is not strictly standardized in actual user-generated texts and loanwords are frequently abbreviated, a set of orthographic variations is observed, especially in nominal MWEs: they need to be recognized and normalized.

In this paper, we introduce the methodology of the construction of the linguistic resource DECO-MWE, in particular, based on a corpus of cosmetics review texts. The procedure used in this study is reproduced in the study of other domain corpora in Feature-based Sentiment Analysis. In Section 2, related work is briefly reviewed. In Section 3, the methodology adopted in this study is described, and in Section 4, four types of MWEs constructed in this study are discussed. A short evaluation of our linguistic resources is presented in Section 5, followed by the conclusion in Section 6.

## 2. Related Work

As a type of MWEs, idiomatic expressions are new units where compositionality is not observed. Since many of them may exhibit polarity values, it is important to take them into account in sentiment analysis (Williams et al. 2015). De Marneffe et al. (2008) point out that the words that constitute MWEs are combined into a single expression with a meaning independent of the individual constituents. In other words, the expression is represented within a speaker's mental lexicon just like a single word (Jackendoff 1997). In addition, the problem of analyzing text semantics can be exacerbated by the scalar variability of the expressions, as the study of Piao et al. (2003) reveals. Therefore, more attention needs to be drawn to computational MWE processing.

Baldwin and Kim (2010) define Compound Nominalization as a combination of two or more nouns into an MWE. In the perspective of feature-based sentiment analysis, named entities or feature nouns can appear in the form of compound nouns, which adds weight to the necessity to process them. Tanaka and Baldwin (2003) and Lapata and Lascarides (2003) concretely examine compound nouns and their components based on the British National Corpus (Burnard, 2000). Tanaka and Baldwin (2003) find that NN compound nouns actually cover 1.44% of the corpus . In most studies, machine learning is applied for MWE processing since it requires significant time and cost to construct sizeable linguistic resources manually.

Taboada et al. (2011) take into account MWE processing for sentiment analysis. Their study is grounded in a careful examination of sentiment linguistic resources, focusing on lexicon-based sentiment analysis of English texts using a Sentiment Orientation Calculator (SO-CAL). However, only a handful of MWEs (177 entries) such as phrasal verbs (*fall apart*) are involved in the analysis. Williams et al. (2015) emphasize the importance of MWEs as sentiment expressions and process them for sentiment analysis by making use of regular expressions to chunk them as a means of tokenization. This study is relevant in that it concentrates on the role of MWE processing in sentiment analysis, but the coverage of MWEs is limited to 580 entries, which are somewhat far from the practical usage in communication, as they have been extracted from an educational website. Besides, the chunking method based on regular expressions does not scale up to a morphologically complex language such as Korean, since the highly complex and accurate regular expressions required to process inflected forms would be difficult to modify and expand.

Most of the studies on processing MWEs for sentiment analysis are mainly conducted for English, with relatively little attention to other languages. Especially, it is difficult to process Korean MWEs because a Korean word should be analyzed into several morphemes, namely, lexical items combined with various inflectional postpositions (Kim and Shin, 2013). It means that some elements of an MWE can be followed by multiple inflectional suffixes such as nominal postpositions (in Korean, *josa*) or verbal/ adjectival postpositions (*eomi*). Moreover, it may be troublesome to capture the boundary of a word because spacing rules are not strictly respected in unstructured texts (Lee 2001).

Considering the limitations of current studies on MWE processing, this paper concentrates on constructing linguistic resources for properly recognizing and extracting Korean MWEs in the FBSA approach.

## 3. Methodology for Construction of DECO-MWE Resources

### 3.1 Data Collection

The rapid growth of Korean cosmetic industry positioned Korea as the tenth biggest market worldwide with its estimated market value of $7,427 million US dollar in 2015 (Kim 2017), leading to an increased demand for fine-grained SA.

To explore sentiment MWEs in cosmetics reviews, we crawled reviews and online User-Generated Contents, and extracted 31,506 cosmetic product names and 468 brand names from a Korean cosmetics review website called *Powder-Room*[1]. The review data is made of 796,689 tokens and 56,354 sentences.

To collect MWEs representing sentiment polarity, we first divided all sentences into two groups: sentences with occurrences of polarity words registered in DECO-Lex (Nam 2015) as sentiment words (QX- tags), and sentences without these sentiment words. This was done by applying DECO-Lex to the corpus with the Unitex platform (Paumier 2003). Then, from the second group of sentences, the most frequent neutral words (i.e. words that are assigned a tag of context-dependent polarity (QXDE) in DECO-Lex entries) were selected by computing the term-frequency table, and their concordances were generated. Our assumption was that the frequent tokens exhibiting no polarity may have a considerable role in the composition of Polarity-MWEs.

For the MWEs of named entities, 31,506 cosmetic product names and 468 brand names were examined to predict the syntagmatic combination of sequences. For the MWEs denoting features, we focused on some types of frequent words such as the equivalents of 'color', 'ingredient', 'scent', selecting them by consulting the sub-menus of the website.

Figure 1 describes how to construct the DECO-MWE resources systematically. After extracting and sorting the MWEs as described above, we utilized the Local Grammar Graph (LGG) formalism (Gross 1997, 1999), represented linguistic patterns in LGGs and compiled the LGGs into Finite-State Transducers (FST) through the Unitex platform (Paumier 2003). There is a coupling between the LGGs and DECO-Lex, as the LGGs use the lexical information stored in the dictionary.

The LGGs, DECO-Lex and DECO-Tagset represent syntactically complex patterns elegantly and enable correct tokenization of morphologically complicated sequences. As mentioned above, these resources associate Korean MWEs with a notably complex set of inflectional variations. Such complexity is a typological property of agglutinative languages. The DECO dictionary provides the information of all possible combinations of nouns and *josa* postpositions as well as those of predicates and *eomi* postpositions. When DECO-Lex is applied to the corpus for morphological analysis, the lexical information registered in DECO-Lex is recognized automatically by LGGs, which makes it possible to tokenize MWEs and normalize them with the canonical description. The benefits of utilizing Local Grammar Graphs may be summarized as follows:

- The flexibility of MWE processing, with the possibility of specifying input in various forms: phonemes, syllables or words.
- Compatibility with DECO-Lex and DECO-Tagset to use part of speech(POS), lemma and surface forms.
- Expressive power for lexico-syntactic irregularities and normalization.

[1] Korean cosmetic review website: www.powderroom.co.kr

### 3.2 Overview

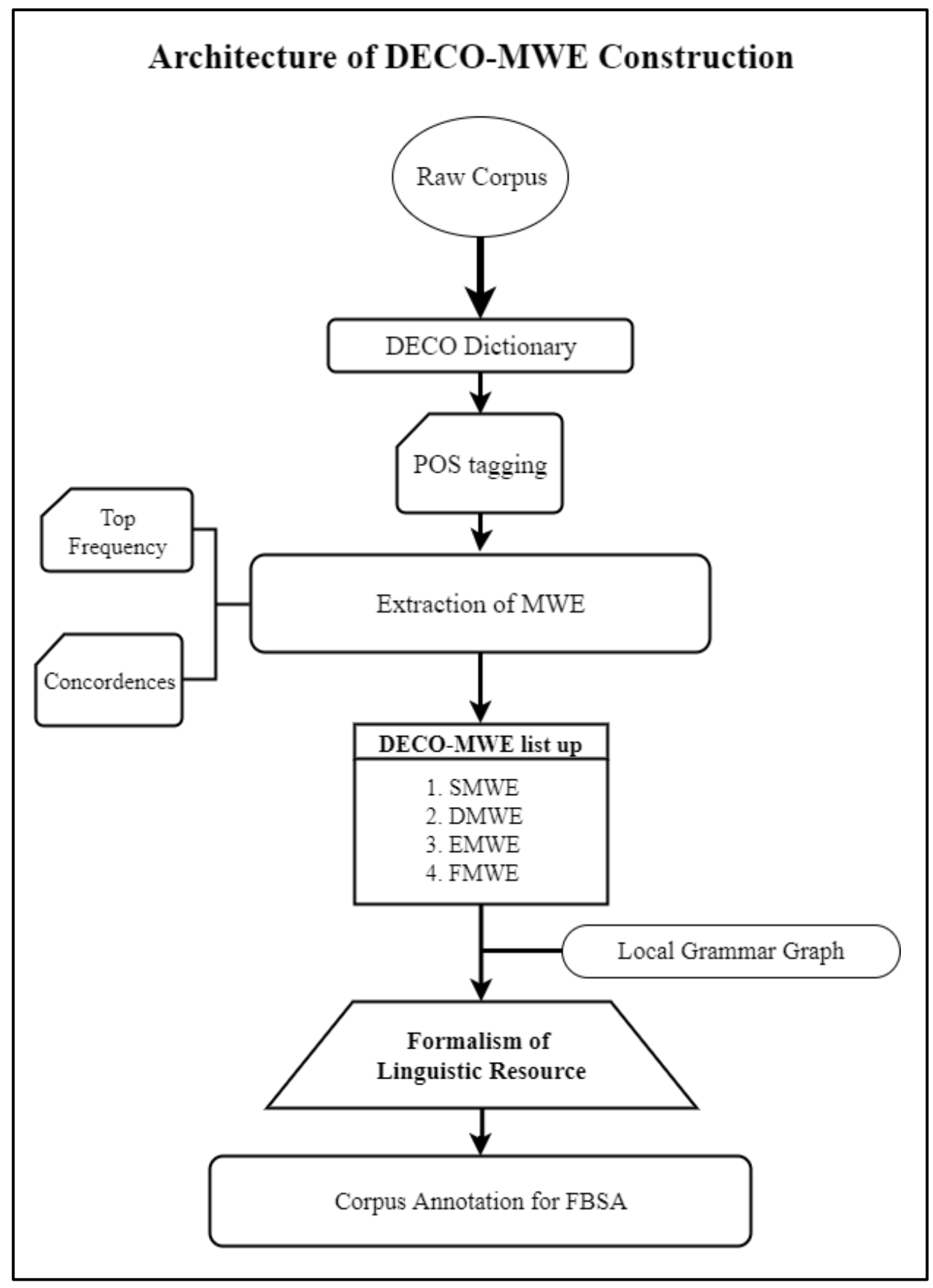

Figure 1: Architecture of DECO-MWE construction

## 4. The DECO-MWE Resources

DECO-MWE covers 4 types of MWEs: Standard Polarity MWEs (SMWEs), Domain-dependent Polarity MWEs (DMWEs), Compound Named Entity MWEs (EMWEs) and Compound Feature MWEs (FMWEs).

### 4.1 Polarity MWEs

Polarity MWEs are the most important keywords of all MWEs for FBSA. Given that MWEs are lexical units that consist of more than one word delimited by white-space (Sag et al., 2002), Polarity MWEs represent polarity expressions with unpredictable meaning. For example, *That's a rip off* means that something is too expensive for what it is, and does not refer to an actual theft. We define Polarity MWEs as MWEs holding polarity values (i.e. expressive positive or negative opinion) in FBSA.

We categorized Polarity MWEs in types according to their syntactic structures. There are three types of combination of nouns and predicates:

- Noun + Noun
- Noun + Predicate

- Predicate + Predicate (including *josa* and *eomi* since each expression can be inflected in various ways)
- ETC (Idiosyncratic sequences)

We constructed LGGs based on the examination of lexical-syntactic patterns observed in the cosmetics domain. Since the interpretation or properties of a significant amount of MWEs vary according to various domains, it seems more effective to divide these MWEs into two sub-categories: Standard Polarity MWE (SMWE) and Domain-dependent Polarity MWE (DMWE).

Notice that the lexicon-based approach to SA aims to tokenize multiword expressions having semantic orientation. Therefore, we assign polarity values in the form of the DECO-PolClass tagset (Nam 2015) and the tokenization is performed by chunking and tagging MWEs with the LGGs.

### 4.1.1 Standard Polarity MWEs (SMWEs)

SMWEs convey a unique polarity value regardless of sub-domains. An example of SMWEs is '바가지를 쓰다/*pakaci-leul sseuta*'(-) (similar to the English expressions *cost an arm and a leg, cost a fortune, pay through the nose*).

In order to construct a sizable quality SMWE resource, we considered not only the 82 SMWEs extracted from the cosmetic corpus but the 205 lists proposed by existing research (Kim, 2000) as well. Additionally, we supplemented the resource by listing up meticulously selected 834 SMWEs from the extensive data crawled from the web-based idiom dictionary [2]. This complementary approach makes the SMWE cover extensive idiomatic MWEs holding polarity values quantitatively as well as qualitatively. The SMWEs were formalized in LGGs.

Once a reliable list of MWEs is prepared, LGGs representing these MWEs are manually constructed under a form of the directed graph as follows:

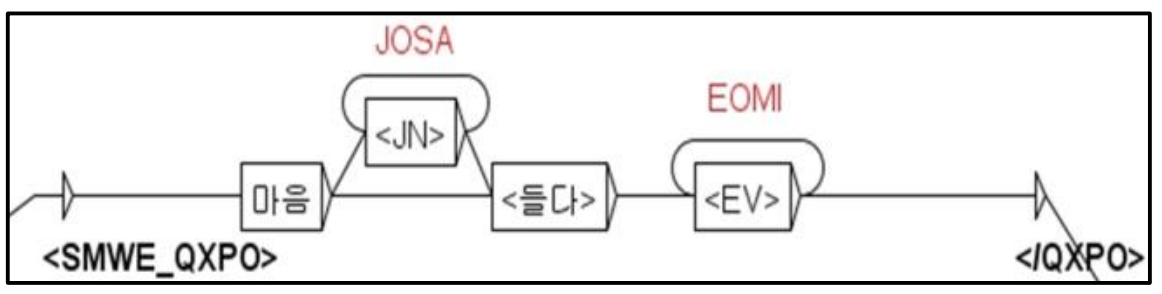

Figure 2. An example of Positive SMWE

The LGG in Figure 2 represents certain MWEs such as '마음에 들다/*maeum-ey teulta* ('to catch one's fancy'). In this LGG, units without '<' and '>' represent surface forms (e.g. 마음), while those with < > such as <들다> represent lemma forms. As a matter of fact, <들다> recognizes a verb root when it is followed by a certain number of inflectional suffixes recognized by the <EV> symbol. By chunking the expressions and enclosing them in XML-like tags such as <SMWE_QXPO> and </QXPO>, this LGG tags automatically SMWEs in accordance with their corresponding semantic orientation.

The overall figures of SMWEs are classified by polarity orientation as shown below.

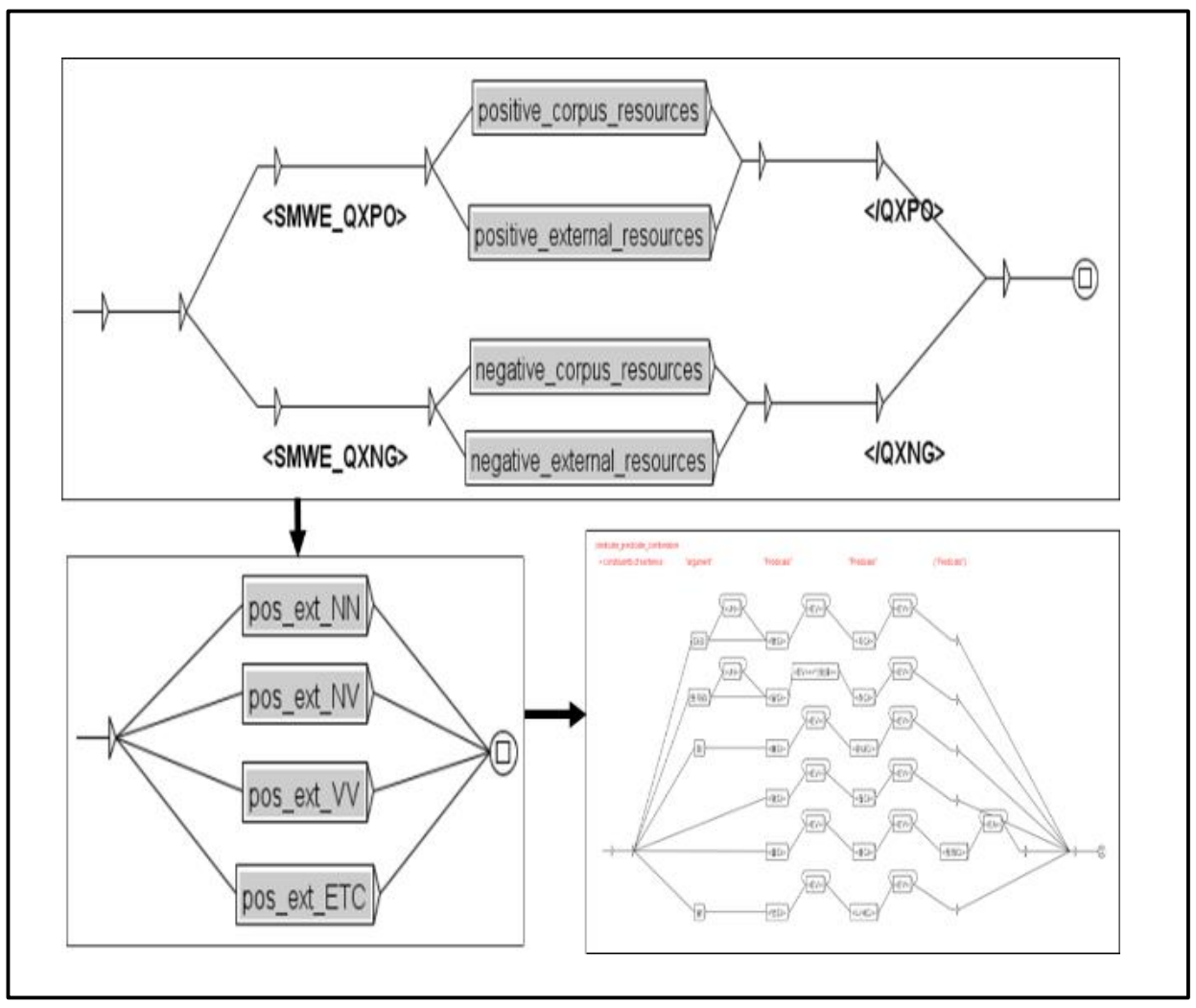

Figure 3: Overall SMWE LGG excerpt

Figure 3 illustrates the main LGG which contains the sub-graphs (e.g. Positive SMWE sub-graphs) to process extensive data of SMWEs. The LGGs representing SMWEs in this study include 1,121 types.

### 4.1.2 Domain-Dependent Polarity MWEs (DMWEs)

Let us consider:

(3a) 촉촉하게 스며들다/*chokchokhagey seumyeo teulta* ("something soaked into skin moistly" (+))

(3b) 모공 부각/*mokong pwukak* ("skin pore expansion"(-))

(3c) 빛이 나다/*pich-i nata'* ("shine on the skin" (+))

Certain Polarity MWEs extracted from the corpus such as (3) convey domain-dependent polarities, thus they are classified as DMWEs: they belong to a specific domain where each MWE shows its own unique meaning. In terms of semantic properties, they may be classified as figurative expressions, having a vital role in conveying positive or negative opinion. Overall, distinguishing SMWEs and DMWEs has a practical advantage in extending polarity MWEs for FBSA. Figure 4 is an example of DMWEs:

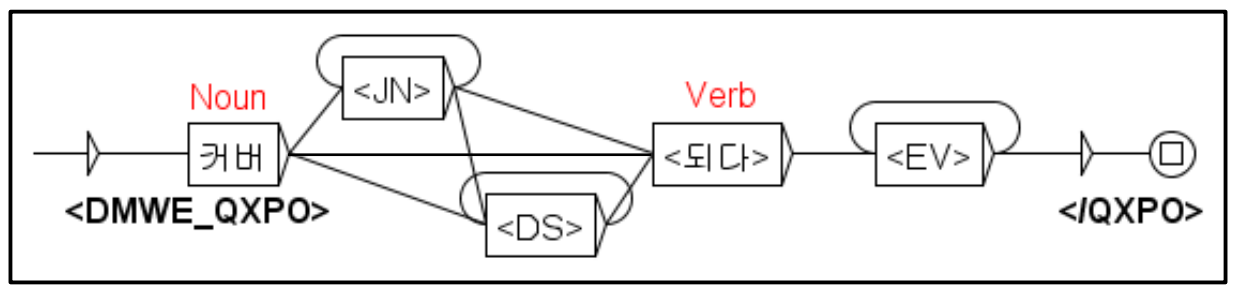

Figure 4: A LGG of Positive DMWE

This LGG represents certain sequences such as '커버가 되다/*kheopeo-ka toyta*'("perfect cover makeup(+)") that conveys a positive opinion without any explicit sentiment word. The verb <되다> is inflected by the postposition recognized by the <EV> symbol, and the noun '커버/*kheopeo*' is followed by one or several postpositions recognized by <JN>. The possibility of adverbial modification is marked by the <DS> symbol recognizing the insertion of possible adverbial words such as '잘/*cal* (well)' or '완전히/*wanceonhi* (completely)'.

[2] NAVER Idiom Dictionary: krdic.naver.com/list.nhn?kind=idiom

The overall figures of DMWEs are sorted on the similar basis of SMWEs classification as depicted below.

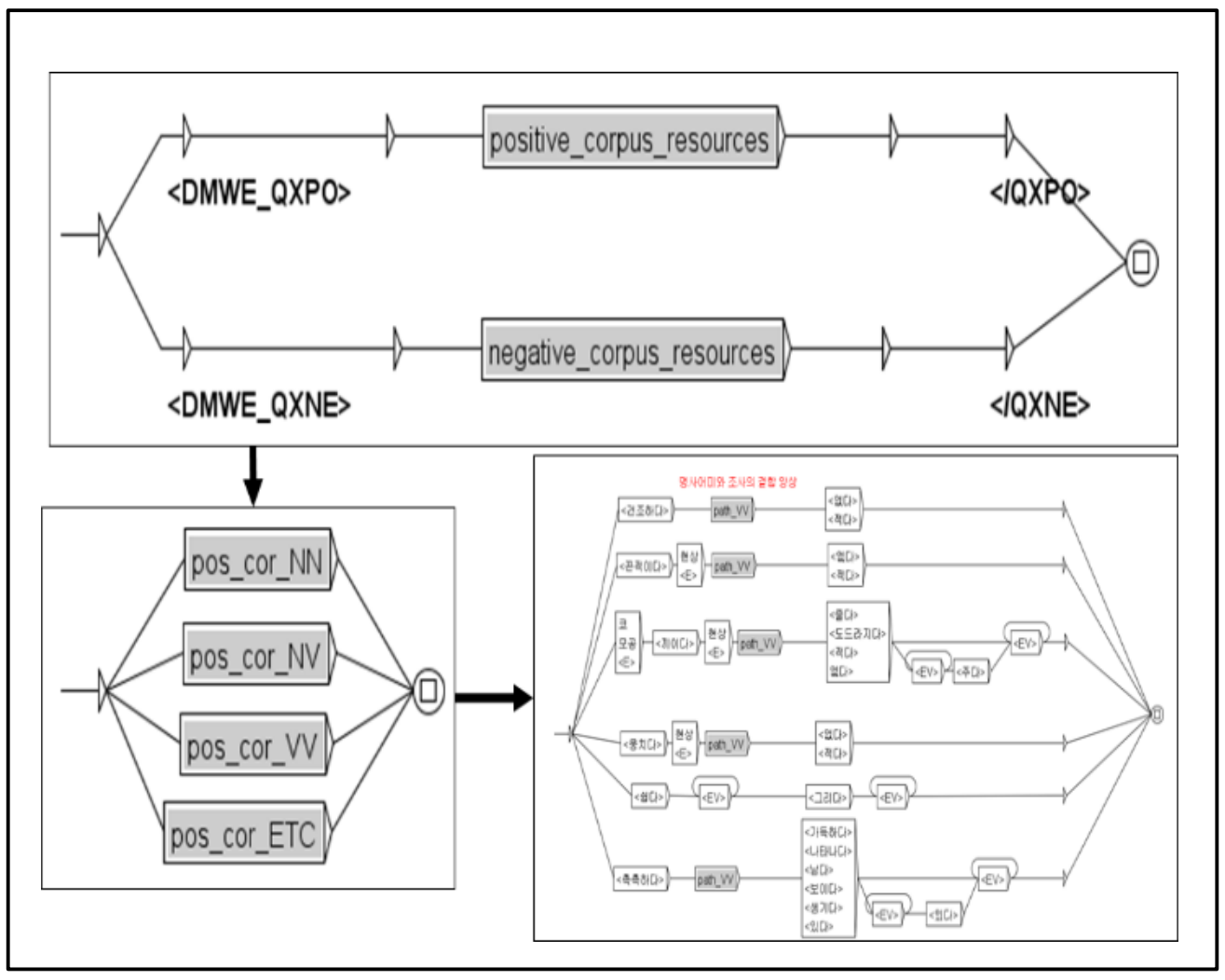

Figure 5: Overall DMWE LGG excerpt

Figure 5 depicts the main LGG with the sub-graphs (e.g. Positive DMWE sub-graphs) analyzing many DMWE sequences. The LGGs can process 1,576 types of DMWEs.

## 4.2 Compound Named Entity & Feature MWEs

In addition to polarity MWEs, complex named entities and feature nouns should also be correctly analyzed in FBSA. In particular, the majority of cosmetic brands and product names are made up of several words and are right-headed, exhibiting 'Modifier-Head' structure. As opposed to Polarity MWEs that may be noun phrases, verbal phrases or adjectival phrases, these MWEs are basically noun phrases. In addition, they are not particularly related to subjective opinion, but to topics. In FBSA, they mostly play a role of Target (e) or Aspect (a) of the opinion or sentiment. Therefore, it will be crucial to properly recognize named entity MWEs and feature noun MWEs to succeed in FBSA. In this study, we divided these MWEs into 2 sub-categories: Named Entity MWEs (EMWEs) and Feature Noun MWEs (FMWEs).

### 4.2.1 Named Entity MWEs (EMWEs)

Based on the term-frequency table of EMWEs, we observe that components may be basically brand names, modifiers, heads (referents) or post-modifiers. Generally, a head denotes the referent of the entity such as 'cream', 'toner', 'foundation' or 'mascara' whereas a modifier represents aspects of its referent such as 'moisture' or 'essential'. In Korean, most EMWEs are borrowed from English words, and therefore variations of transliteration weaken recall in automatic recognition of these terms. In addition to these morpho-phonological variations, elision or contraction of certain units occurs in user-generated review texts. These irregularities can be legibly and successfully controlled with the LGG formalism. In the LGG in Figure 6, the combinations of the variable elements are described in a finite-state way. The application of the LGG delimits and normalizes the combinations with the XML-like tags <EMWE-XXPR> and </XXPR>. ('XXPR' annotates a sub-category of Named Entity registered in DECO-Lex: 'Product/Brand Name').

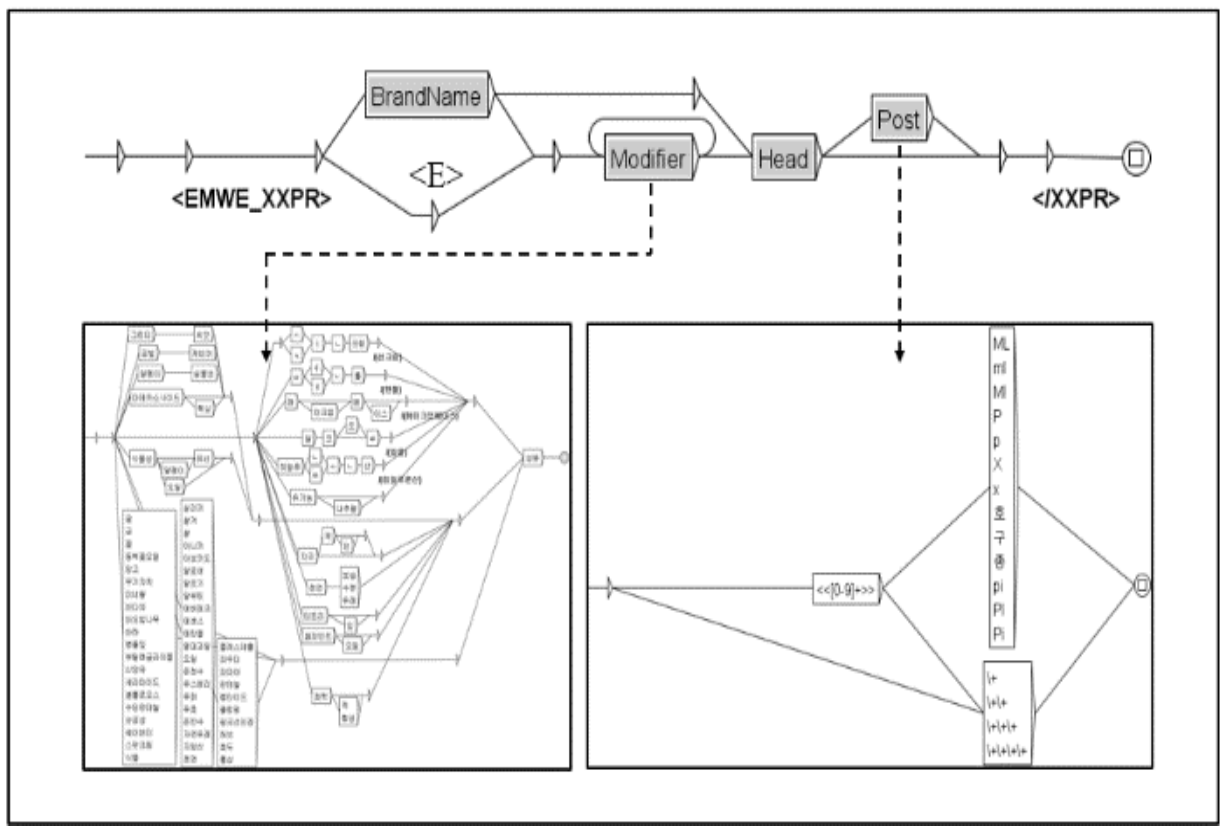

Figure 6: An example of LGGs for EMWE

The LGG in Figure 6 represents the EMWEs consisting of combinations of a brand name, modifiers, a head and a post-modifier. The grey boxes call the sub-graphs. The <E> path makes the brand name optional, which covers EMWEs made up of a product name. The Brand Name LGG is a sub-graph which contains the 468 units of cosmetic brand names observed in the cosmetic review website. Other sub-graphs represent the aggregation of multiple LGGs that recognize 31,560 product names. The LGG in Figure 6 not only includes the lists of products but also retrieves diverse variations caused by elision or contraction of the most frequent tokens. This LGG chunks EMWEs and assigns them a category by the XML-like tags <EMWE_XXPR> and </XXPR>.

The LGG in the left bottom of Figure 6 displays a part of the Modifier LGG: it recognizes modifiers most frequently collocating with a referent. The modifiers denote certain features of the referent, including function and substance. Let us consider:

(4a) 헤라 셀/ *heyla seyl* ('Hera(a brand name) Cell')

(4b) 헤라 셀 에센스/ *heyla seyl eyseynseu* ('Hera Cell (a modifier) Essence')

(4c) 헤라 에센스/ *heyla eyseynseu* ('Hera Essence(a referent)')

(4d) 셀 에센스/ *seyl eyseynseu* ('Cell Essence')

All these examples are legibly treated in the form of LGGs, since the combinatorial properties of each element are directly represented by finite-state transducers. In this way, a great number of complex modifiers may be formalized in LGGs. Predictable complex types may be added in this LGG, even if they are not observed in the corpus.

As a result, EMWEs formalized by LGGs in this study include around 31,560 types.

### 4.2.2 Feature Noun MWEs (FMWEs)

EMWEs show a relatively small range of phonological variation since many are proper names for which local partners have already chosen a transliteration. In contrast, FMWEs show much more variations since most are English common nouns that users can choose how to transliterate into Korean. Thus, it is outstandingly beneficial to consider their morpho-phonological variations in order to process them properly.

Let us consider the example of 'color' that is one of the most frequent feature nouns in this domain. In Korean, this term, as an English loanword, may occur under several forms due to vowel and consonant variations. Consider:

(5a) 컬러감/ *kheolleo-kam* ('color-feeling=color')

(5b) 칼라감/ *khalla-kam* ('color-feeling=color')

(5c) 칼라 정도/ *khalla-ceongto* ('color-degree=color')

(5d) 컬러밝기/ *kheolleo-palkki* ('color-brightness=color')

The LGGs representing Feature (a) of product manage to cover the whole case of combinable types including a series of strings in several units of word whether they include white-space or not. These variations can be controlled by the following LGG (Figure 7).

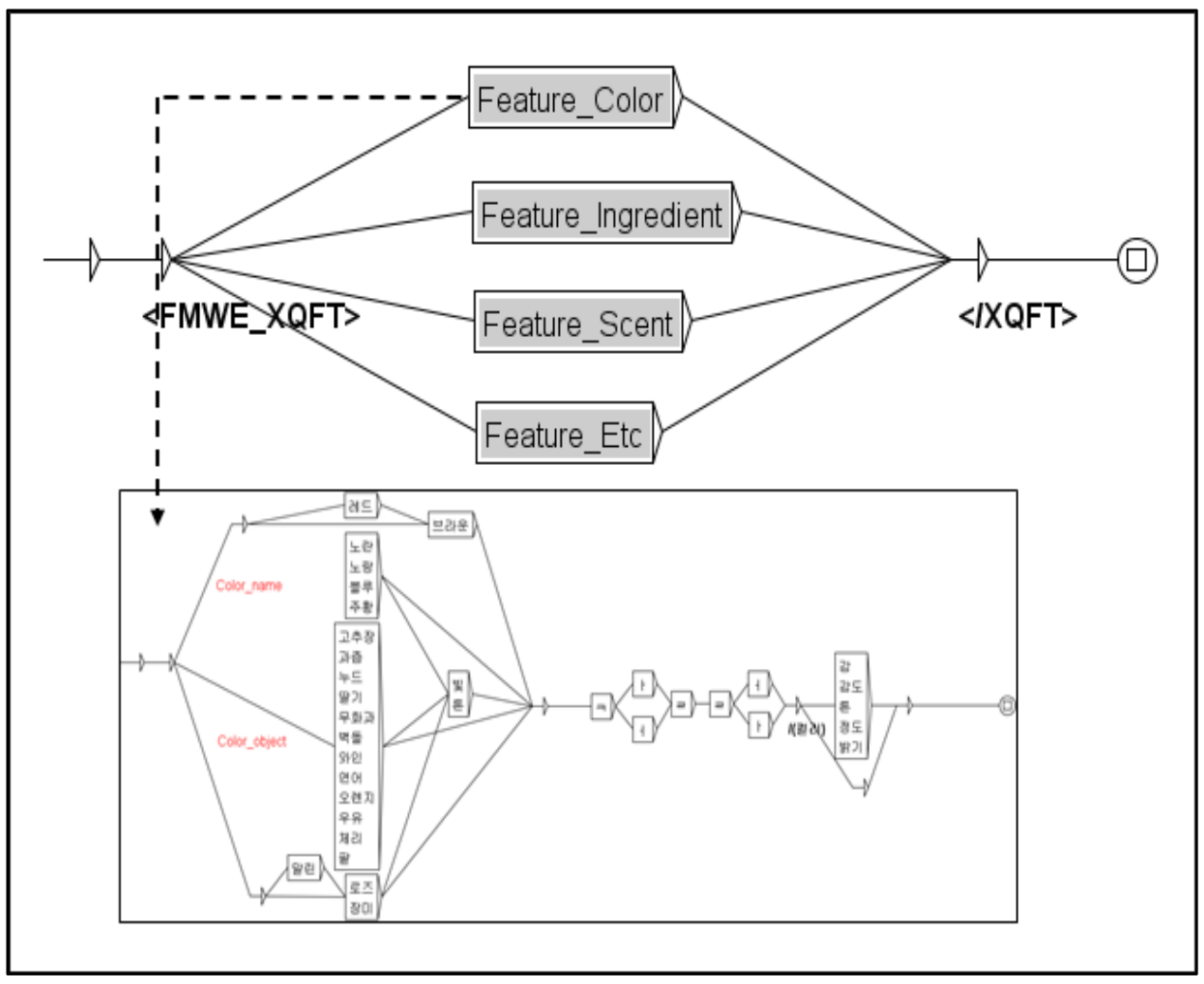

Figure 7: An example of an LGG for FMWE

The sub-graph included in the LGG on Figure 7 is organized in the form of word parts connected together and describes complex feature nouns such as '오렌지 칼라/*oleynci khalla* (orange color)' or '블루 컬러감/ *peullwu kheulleu-kam* (blue color-feeling = blue color)'.

We formulate the FMWE composition by frequent collocations of Feature's headwords, grounded in the term-frequency table.

These FMWEs are chunked and assigned a category with tags which will be crucial to normalizing these variations, such as <FMWE_XQFT> and </XQFT>.

In this study, FMWEs processed by LGGs involve around 165 types.

## 5. Evaluation

In order to evaluate the linguistic resources proposed in this study, we requested thirty cosmetics reviewers to build a test corpus for the performance evaluation of our resources. The corpus consists of 5,870 tokens(300 sentences) and contains several polarity MWEs and compound noun MWEs as follows:

| Polarity MWE | | CompoundN MWE | | Total |
|---|---|---|---|---|
| SMWE | DMWE | EMWE | FMWE | 427 |
| 36 | 79 | 266 | 46 | |

Table 1: Number of MWEs in the test corpus

Three researchers who majored in linguistics were responsible for the labor-intensive annotation to tag MWE on the test corpus. They cross-checked the tagged corpus based on the strict inter-annotator agreement to be fully served as the evaluation criteria.

We compared the result automatically obtained by the application of LGGs into this corpus with the manually detected result in Table 1. Table 2 shows the result of this evaluation:

| | SMWE | DMWE | EMWE | FMWE | Total |
|---|---|---|---|---|---|
| Precision | 0.933 | 0.936 | 0.797 | 0.948 | 0.845 |
| Recall | 0.777 | 0.746 | 0.770 | 0.804 | 0.770 |
| F-Measure | 0.848 | 0.830 | 0.783 | 0.870 | 0.806 |

Table 2: Performance evaluation

The F-measure turns out to be 0.806 while recall is 0.770 and precision 0.845.

As the result, the overall recall shows lower than the precision, but it seems similar to precision (0.797) and recall (0.783) in the case of the EMWEs. Thus, it caused the F-Measure of EMWE to be lower than other types. The main reason for this result is attributed to syntactic ambiguity. In the case of a sentence like '왠지 모르게 언니 마스카라가 더 좋더라구요/*waynci molukey enni masukhala-ka te cohtelakuyo* (For some reason, I like a sister mascara)', the phrase '언니 마스카라/*enni masukhala* (sister mascara)' is ambiguous to be analyzed by two approaches. One way is to parse '[$_{np}$ [$_{n}$ 언니] [$_{n}$ 마스카라]]/*enni masukhala* (a sister mascara)' as an EMWE which refers to a 'brand'(*enni*) mascara, and the other way is to analyze '[$_{np}$ [$_{np}$ [$_{n}$ 언니][$_{pos}$ (의)]] [$_{n}$ 마스카라]]/*enni(uy) masukhala-ka* (sister's mascara)' caused by ellipsis in noun phrases with possessive case: Korean genitive *josa* '의/*uy*', meaning 'a sister's mascara'. Such linguistic ambiguity caused by Korean *josa* omission carries difficulty with recognizing EMWEs for the accurate result.

## 6. Conclusion

This paper presents a linguistic resource of Korean Multiword Expressions for Feature-Based Sentiment Analysis (FBSA): DECO-MWE. To construct linguistic resources of sentiment MWEs efficiently, we utilized the Local Grammar Graph (LGG) methodology: DECO-MWE is formalized as a Finite-State Transducer that represents lexical-syntactic restrictions on MWEs.

In this study, we built a corpus of cosmetics review texts, which show particularly frequent occurrences of MWEs. Based on the empirical examination of the corpus, four types of MWEs have been discerned. The DECO-MWE thus covers the following four categories: Standard Polarity MWEs (SMWEs), Domain-Dependent Polarity MWEs (DMWEs), Compound Named Entity MWEs (EMWEs) and Compound Feature MWEs (FMWEs). The retrieval performance of the DECO-MWE shows 0.806 f-measure in the test corpus.

This study brings a two-fold outcome: first, a sizeable general-purpose polarity MWE lexicon, which may be broadly used in FBSA; second, a finite-state methodology adopted in this study to treat domain-dependent MWEs such as idiosyncratic polarity expressions, named entity expressions or feature expressions, and which may be reused in describing linguistic properties of other domains.

## 7. Bibliographical References